\documentclass{article}

\usepackage[nonatbib, final]{neurips_2021}

\usepackage[utf8]{inputenc} 
\usepackage[T1]{fontenc}    
\usepackage{hyperref}       
\hypersetup{colorlinks,allcolors=black}
\usepackage{url}            
\usepackage{booktabs}       
\usepackage{amsfonts}       
\usepackage{nicefrac}       
\usepackage{microtype}      
\usepackage{xcolor}         
\usepackage{graphicx}
\usepackage{amsmath, amssymb}
\usepackage{multirow}
\usepackage{array}

\newcommand{\norm}[1]{\left\lVert#1\right\rVert_2}

\usepackage{biblatex} 
\title{Fourier-Based Augmentations for Improved Robustness and Uncertainty Calibration}

\author{%
  Ryan Soklaski\thanks{
DISTRIBUTION STATEMENT A. Approved for public release. Distribution is unlimited.
This material is based upon work supported by the Under Secretary of Defense for Research and Engineering under Air Force Contract No. FA8702-15-D-0001. Any opinions, findings, conclusions or recommendations expressed in this material are those of the author(s) and do not necessarily reflect the views of the Under Secretary of Defense for Research and Engineering.
© 2021 Massachusetts Institute of Technology.
Delivered to the U.S. Government with Unlimited Rights, as defined in DFARS Part 252.227-7013 or 7014 (Feb 2014). Notwithstanding any copyright notice, U.S. Government rights in this work are defined by DFARS 252.227-7013 or DFARS 252.227-7014 as detailed above. Use of this work other than as specifically authorized by the U.S. Government may violate any copyrights that exist in this work.}\\
  MIT Lincoln Laboratory \\
  Lexington, MA 02421-6426 \\
  \texttt{ryan.soklaski@ll.mit.edu} \\
\And
  Michael Yee\\
  MIT Lincoln Laboratory \\
  Lexington, MA 02421-6426 \\
  \texttt{myee@ll.mit.edu} \\
\And
  Theodoros Tsiligkaridis \\
  MIT Lincoln Laboratory \\
  Lexington, MA 02421-6426 \\
  \texttt{ttsili@ll.mit.edu} \\
}

\begin{document}

\maketitle

\begin{abstract}
  Diverse data augmentation strategies are a natural approach to improving robustness in computer vision models against unforeseen shifts in data distribution. However, the ability to tailor such strategies to inoculate a model against specific classes of corruptions or attacks---without incurring substantial losses in robustness against other classes of corruptions---remains elusive. In this work, we successfully harden a model against Fourier-based attacks, while producing superior-to-\texttt{AugMix} accuracy and calibration results on both the CIFAR-10-C and CIFAR-100-C datasets; classification error is reduced by over ten percentage points for some high-severity noise and digital-type corruptions. We achieve this by incorporating Fourier-basis perturbations in the \texttt{AugMix} image-augmentation framework. Thus we demonstrate that the \texttt{AugMix} framework can be tailored to effectively target particular distribution shifts, while boosting overall model robustness.
\end{abstract}

\section{Introduction}

Despite the chart-topping performances of CNN-based models across both standard and domain-specialized computer vision benchmarks, the tendency for these models to behave unreliably in the face of subtle changes to data distribution make them likely to fail in deployment \cite{geirhos2018generalisation,globerson2006nightmare,moosavi2017universal}. This lack of robustness has been broadly attributed to the proclivity for CNNs to fit on unintuitive and superficial patterns that exist in the training data \cite{geirhos2018generalisation,jacobsen2018excessive,jo2017measuring}. Thus there is a need  for more comprehensive benchmarks---including those that test models against common corruptions and perturbations \cite{he2021towards,hendrycks2018benchmarking,hendrycks2021natural,lu2020harder,mu2019mnist,recht2018cifar,recht2019imagenet}---as well as for more varied data augmentation strategies.

An analysis of these common image corruptions from a Fourier perspective characterized the degree to which natural perturbations can be summarized by high, mid, and low-frequency variations in intensity over pixel-space \cite{yin2019fourier}. It demonstrated that performance trade-offs in robustness between different categories of corruptions often manifest across different characteristic frequency regimes, e.g., a targeted improvement in robustness to Gaussian noise (high-frequency characteristics) will be concomitant with marked loss in robustness to a fog corruption (low-frequency characteristics) \cite{ford2019adversarial,yin2019fourier}. Whereas targeted data-augmentation strategies suffer from this trade-off, those strategies that mix a diverse set of augmentation primitives--- including low, mid, and high-frequency characteristics---have proven to be effective at boosting overall model robustness~\cite{cubuk2018autoaugment,hendrycks2019augmix}; the \texttt{AugMix} augmentation method is particularly outstanding in this regard \cite{hendrycks2019augmix}.

Still outstanding is the need to build a model with robustness to particular distribution shifts---in a targeted manner---while maintaining robustness to common corruptions. There are well-known universal adversarial perturbations (UAP) that can be applied uniformly across images in order to induce high error rates across convolution-based computer vision models \cite{moosavi2017universal}. Indeed, it has been shown that even simple sinusoidal noise, designed to target particular frequency-and-orientation regimes, can serve as an effective UAP; these so-called Fourier-basis attacks can reduce image-classification models to near-guessing performance \cite{hendrycks2019augmix,tsuzuku2019structural,yin2019fourier}. It is of high importance that models can be made robust both to common, naturally occurring corruptions, as well as known black-box adversarial perturbations such as UAPs.

In this work we investigate the capacity for \texttt{AugMix} to be modified to target particular data distribution shifts, while maintaining its outstanding ability to instill robustness against common corruptions. We specifically seek protection against Fourier-basis attacks, which are noteworthy for their simplicity, their efficacy at diminishing model performance, as well as their ability to be tailored to target any frequency regime in input space.

\begin{figure}[h]
  \centering
  \includegraphics[width=5.5in]{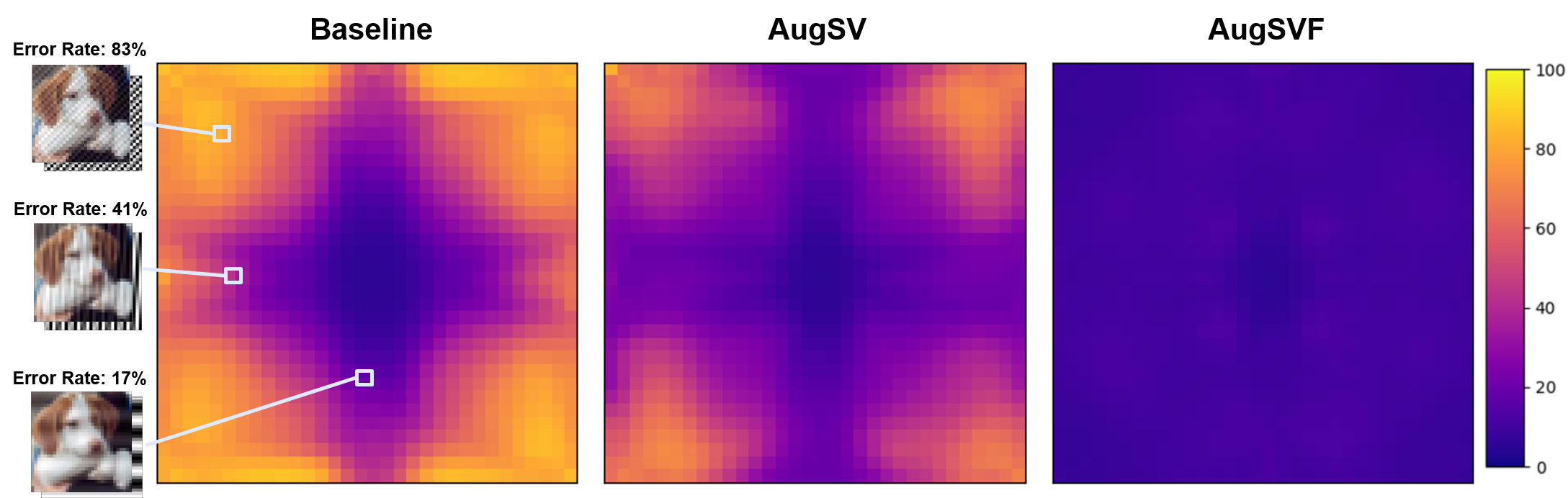}
  \caption{\small Classification error heatmaps for a model trained using three data augmentation techniques, in association with Fourier-basis perturbations applied to all CIFAR-10 test images.
  \texttt{AugSV} (i.e, \texttt{AugMix} \cite{hendrycks2019augmix}) improves robustness to some high and mid-frequency regions. \texttt{AugSVF}, which explicitly incorporates distinct Fourier-basis augmentations, greatly reduces the model's susceptibility across all frequency regimes.}
  \label{fig:heatmaps}
\end{figure}

\section{Fourier-Basis Perturbations} \label{sec:fourier}


A real-valued 2D Fourier-basis matrix of shape $d_x \times d_y$, $U(\vec{k}) \in \mathbb{R}^{d_x \times d_y}$, has its elements defined by a 2D sinusoidal plane wave that is evaluated on the discrete grid $P = \mathbb{I}^{\{0, ..., d_x-1\}\times\{0, ..., d_y-1\}}$, i.e., 
\begin{equation}\label{eq:planewave}
U_{x,y}(\vec{k}) = A \cos (2 \pi f \vec{r} \cdot \hat{k} - \phi)
\end{equation}
 where $\vec{r} = [x, y] \in P$. $A$ is selected such that $\norm{U}$ is fixed.\footnote{The definition of a 2D Fourier-basis matrix provided in section 2 of \cite{yin2019fourier} is written in terms of the behavior of $U_{x,y}$ under the action of a 2D discrete Fourier transform. While their definition may not immediately resemble that of Equation \ref{eq:planewave}, the two are in fact identical.} Each basis matrix is characterized by a so-called wave vector, $\vec{k} = [k_x, k_y] \in K \subseteq P$, whose magnitude determines the frequency of the sinusoid, $f = \norm{(k_x/d_x, k_y/d_y)}$, and whose direction determines the orientation of the wave. $K$ ($\subseteq P$) corresponds to the set of non-degenerate plane waves with a common phase shift of $\phi = \frac{\pi}{4}$.\footnote{The cardinality of $K$ has an upper bound of $d_1 \times (\left \lfloor{\frac{d_2}{2}}\right \rfloor +1)$ \cite{newman_2013}.}
Three example basis matrices are depicted in Figure \ref{fig:heatmaps}.

A Fourier-basis perturbation consists of adding $U(\vec{k})$, elementwise, to each color channel of a shape-${d_x \times d_y}$ image, with strength controlled by $\norm{U}$. The perturbation can also be applied to each channel with a randomly-drawn sign applied to it; this is referred to as a ``random-flip'' perturbation.

Susceptibility heatmaps for channel-aligned perturbations with $\norm{U} = 2$ are shown in Figure \ref{fig:heatmaps}. It reports $d_x \times d_y$ metric values that measure a model's performance on $d_x \times d_y$ perturbed test sets. Entry $[k_x, k_y]$ ($\in K$) corresponds to a perturbation using $U(\vec{k} = [k_x, k_y])$ applied identically to every image in the test set; the color of the pixel at $[k_x, k_y]$ thus conveys the model's classification error rate on that perturbed test set. The heatmap is arranged so the lowest-frequency resides at its center; it is symmetrized about its center, due to the noted degeneracies among $\vec{k} \in P$.

\subsection{Incorporating Fourier-Basis Perturbations in \texttt{AugMix}}

The standard \texttt{AugMix} implementation involves a set of five spatial primitives ($S$) $p_\mathrm{rotate}$, $p_\mathrm{shear_{\{x,y\}}}$, $p_\mathrm{trans_{\{x,y\}}}$, and a set of four ``vision'' primitives ($V$) $p_\mathrm{contrast}$, $p_\mathrm{EQ}$, $p_\mathrm{poster}$, $p_\mathrm{solar}$. \texttt{AugS}, \texttt{AugV}, and \texttt{AugSV} will indicate the inclusion of the primitives in $S$, $V$, and $S \cup V$ in the \texttt{AugMix} framework, respectively. Thus \texttt{AugSV} corresponds to the original implementation in \cite{hendrycks2019augmix} (see Appendix for more details). We can naturally include a Fourier-basis perturbation primitive (denoted by $F$) among the primitives used by \texttt{AugMix}. The elements of stochasticity for this augmentation result from sampling: $\vec{k}$, $\phi$, and $\norm{U}$. We sample over frequencies ($\lVert\vec{k}\rVert_2$) uniformly.\footnote{An unweighted sampling of the Fourier-bases would favor high-frequency perturbations.} For example, \texttt{AugSVF} includes spatial, vision, and Fourier primitives. Lastly, there are four primitives that overlap with the CIFAR-C corruptions, $p_{\mathrm{brightness}}$, $p_{\mathrm{color}}$, $p_{\mathrm{contrast}}$, and $p_{\mathrm{sharp}}$, which we will denote as $C$.

\section{Experimental Results}

We train a Wide ResNet architecture \cite{zagoruyko2016wide} on both CIFAR-10 and CIFAR-100, respectively, using a variety of data-augmentation strategies. We evaluate the following augmentation strategies: \texttt{Base}, \texttt{AugF}, \texttt{AugS}, \texttt{AugSF}, \texttt{AugSV}, \texttt{AugSVC}, and \texttt{AugSVF}. For each augmentation technique, we train the model using four independent random seeds and report the average performance. The baseline augmentation entails randomly flipping and cropping an image, and then normalizing it. We evaluate the model's classification error as well as the RMS calibration error \cite{kumar2019verified}. To assess the robustness of these models to unforeseen, common image corruptions, we include evaluations against all categories and severities of corruptions in the CIFAR-10-C and CIFAR-100-C datasets \cite{hendrycks2018benchmarking}. Our architecture, hyperparameters, and training methods exactly match those used in \cite{hendrycks2019augmix}; all models are trained using Jensen-Shannon divergence as a consistency loss \cite{wiki:jsdloss}.

\begin{table}
\centering
\caption{\small A comparison of augmentation strategies across datasets and performance metrics. CIFAR-C results are averaged across all corruptions and severities. \texttt{AugSVF} produces superior calibration metrics across the board, and yields outstanding robustness on corruption datasets.}
\resizebox{12cm}{!}{%
\begin{tabular}{llrrrrrrr}
\toprule
{} & {Augmentations} & {\texttt{Base}} & {\texttt{AugF}} & {\texttt{AugS}} & {\texttt{AugSF}} & {\texttt{AugSV}} & {\texttt{AugSVC}} & {\texttt{AugSVF}} \\
{Datasets} & {Error Metrics} & {} & {} & {} & {} & {} & {} & {} \\
\midrule
\multirow[c]{2}{*}{CIFAR-10} & Classification & 5.7 & 6.3 & 5.6 & 5.9 & 5.1 & \textbf{5.1} & 5.2 \\
 & RMS Calibration & 7.4 & 7.2 & 5.4 & 5.4 & 3.6 & 4.2 & \textbf{3.3} \\
\midrule
\multirow[c]{2}{*}{CIFAR-10-C} & Classification & 27.0 & 18.1 & 15.2 & 11.6 & 10.9 & 10.9 & \textbf{9.9} \\
 & RMS Calibration & 23.6 & 16.0 & 12.2 & 9.8 & 7.6 & 8.4 & \textbf{6.7} \\
\midrule
\midrule
\multirow[c]{2}{*}{CIFAR-100} & Classification & 25.5 & 27.8 & 26.7 & 27.5 & 24.2 & \textbf{23.5} & 24.9 \\
 & RMS Calibration & 15.9 & 14.7 & 12.6 & 12.4 & 7.4 & 7.9 & \textbf{7.0} \\
\midrule
\multirow[c]{2}{*}{CIFAR-100-C} & Classification & 53.6 & 46.4 & 42.0 & 38.6 & 36.3 & 35.5 & \textbf{34.8} \\
 & RMS Calibration & 33.7 & 25.2 & 21.3 & 18.0 & 13.8 & 14.5 & \textbf{12.1} \\
\bottomrule
\end{tabular}%
}
\label{table:errormetrics}
\end{table}

\subsection{CIFAR-C Evaluations}

The performance metrics reported in Table \ref{table:errormetrics} reveal the efficacy of incorporating Fourier-basis perturbations in the \texttt{AugMix} framework. \texttt{AugSVF} produces the best classification performance on the CIFAR-C datasets and superior calibration results on all benchmarks. The inclusion of \texttt{AugSVC} helps to demonstrate that simply adding any additional primitives to \texttt{AugMix} does not necessarily net a gain in performance. Indeed, \texttt{AugSVC} harms calibration across the board and is inconsistent in its impact on classification error. This is despite the fact that \texttt{AugSVC} explicitly incorporates corruptions that overlap with CIFAR-C.

\begin{figure}[h]
  \centering
  \includegraphics[width=5.5in]{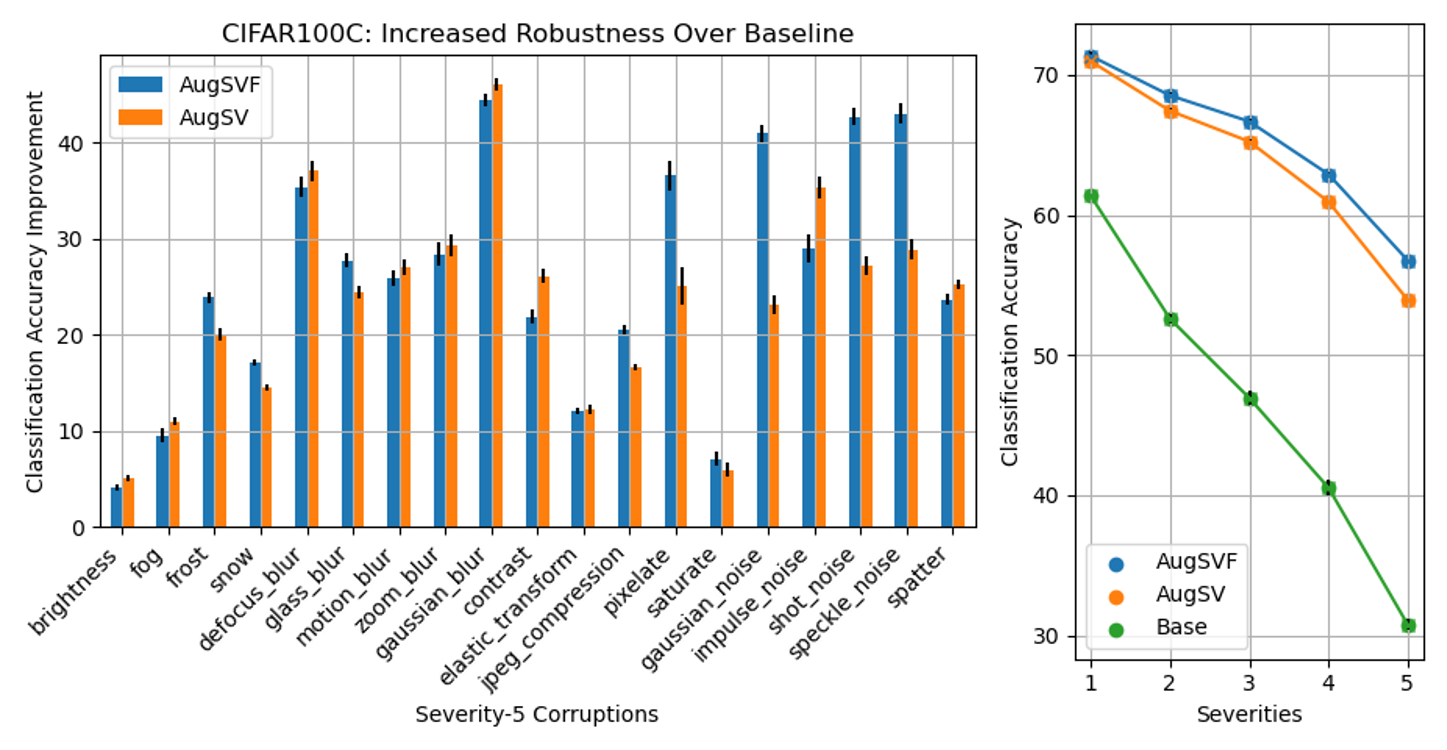}
  \caption{\small (Left) Enhanced robustness in classification accuracy over the baseline, across severity-5 corruption categories for CIFAR-100-C. AugSVF yields a significant boost in robustness to ``noise'' and ``digital'' corruptions, without any significant trade-offs elsewhere. (Right) CIFAR-100-C classification accuracies averaged over corruptions, for each corruption severity.}
\label{fig:c100_bars}
\end{figure}

Examining these results prior to aggregation over severity and corruption-type reveals that the gains in robustness achieved by \texttt{AugSVF} grows relative to \texttt{AugMix} with increasing corruption severity. Figure \ref{fig:c100_bars} shows, for each severity-5 corruption in CIFAR-100-C, the improvement that \texttt{AugMix} and \texttt{AugSVF} achieve over the baseline's classification accuracy for that category. \texttt{AugSVF} offers substantial boosts in robustness for several noise-type and digital-type corruptions, with improvements exceeding ten percentage points for multiple categories. Despite these marked improvements against distinctly ``high-frequency'' corruptions \cite{yin2019fourier}, \texttt{AugSVF} does not suffer any substantial trade-offs in other categories, other than against contrast and impulse noise; it otherwise sustains the gains in robustness achieved by \texttt{AugMix}. The error bars denote standard deviations across independent seeds.

\begin{table}[t]
\centering
\caption{\small Comparing vulnerabilities to Fourier-based attacks. \texttt{AugSVF} greatly reduces the impact that Fourier-basis corruptions have on the model's performance. The phase shift used for all perturbations was held out at train time for \texttt{AugSVF}, nor did \texttt{AugSVF} train on $\norm{U}$ exceeding 2.}
\resizebox{12cm}{!}{
\begin{tabular}{llrrrrrr}
\toprule
{} & {Metric} & \multicolumn{3}{r}{Mean (Max) Class Error} & \multicolumn{3}{r}{Mean (Max) RMS Cal Error} \\
\cmidrule (r){3-8}
{} & {Method} & {\texttt{Base}} & {\texttt{AugSV}} & {\texttt{AugSVF}} & {\texttt{Base}} & {\texttt{AugSV}} & {\texttt{AugSVF}} \\
\cmidrule (r){3-5}
\cmidrule (r){6-8}
{Dataset} & {$||U||_2$} & {} & {} & {} & {} & {} & {} \\
\midrule
\multirow[c]{4}{*}{CIFAR-10} & 1 & 24 (83) & 12 (38) & \textbf{6 (8)} & 22 (70) & 8 (28) & \textbf{5 (6)} \\
 & 2 & 46 (88) & 28 (83) & \textbf{9 (15)} & 38 (86) & 19 (73) & \textbf{6 (11)} \\
 & 3 & 60 (90) & 42 (89) & \textbf{13 (27)} & 48 (89) & 29 (79) & \textbf{9 (20)} \\
 & 4 & 68 (90) & 53 (90) & \textbf{19 (44)} & 54 (90) & 37 (78) & \textbf{13 (32)} \\
\midrule
\multirow[c]{4}{*}{CIFAR-100} & 1 & 55 (92) & 39 (69) & \textbf{27 (29)} & 34 (66) & 16 (43) & \textbf{8 (10)} \\
 & 2 & 74 (98) & 58 (95) & \textbf{31 (40)} & 50 (91) & 30 (82) & \textbf{11 (16)} \\
 & 3 & 83 (99) & 70 (98) & \textbf{36 (55)} & 58 (97) & 43 (94) & \textbf{14 (26)} \\
 & 4 & 88 (99) & 78 (99) & \textbf{43 (69)} & 62 (98) & 52 (96) & \textbf{17 (38)} \\
\bottomrule
\end{tabular}%
}
\label{table:fourier}
\end{table}

\subsection{Evaluating Against Fourier-Based Attacks}

We assess the ability of models that leverage the \texttt{Base}, \texttt{AugSV}, and \texttt{AugSVF} augmentation strategies, respectively, to generalize to data that has been corrupted with Fourier-basis noise. We corrupt the entire test using every Fourier-basis matrix; thus we evaluate all possible frequencies, orientations, and channel-alignments.
The perturbations use a phase shift held out from training for \texttt{AugSVF}, nor did \texttt{AugSVF} include perturbations with $\norm{U} > 2$ during training. 
Table \ref{table:fourier} reports results for both random and targeted Fourier attacks of varying strengths, as measured via mean and max error rates, respectively. The maximum error metrics correspond to a targeted attack where a single, worst-case Fourier-basis is selected to modify the entire test set.

Figure \ref{fig:heatmaps} and Table \ref{table:fourier} report that \texttt{AugSVF} is highly robust to corruptions of all frequencies, whereas \texttt{Base} and \texttt{AugSV} are reduced to near-guessing for targeted attacks with $\norm{U} \geq 2$, and a random $\norm{U} = 1$ corruption roughly doubles their clean error rates. Thus incorporating Fourier perturbations in \texttt{AugMix} is effective at targeting this distribution shift.

\section{Conclusion}
In this work we modify \texttt{AugMix} to protect against Fourier-basis attacks---a particular form of data distribution shift that can be used as an effective universal adversarial perturbation---while improving robustness against common corruptions and uncertainty calibration. We find that inoculating a model against such a fundamental class of perturbations does not degrade performance, despite the noted propensity for models to ``attach'' to spurious statistics. Furthermore, although these perturbations explicitly span all frequencies, there is only an enhancement---and no loss---in robustness to a wider variety of corruptions. Indeed, we obtain outstanding results on CIFAR-10-C and CIFAR-100-C, while greatly ameliorating the model's susceptibility to Fourier-based attacks.

Future work will investigate whether achieving robustness to Fourier-basis attacks provides robustness against other universal adversarial perturbations that have distinctive frequency characteristics. For instance, the UAP developed by Moosavi-Dezfooli et al. \cite{moosavi2017universal} has been shown to exploit narrow regimes of high-frequency patterns to fool models \cite{tsuzuku2019structural}; however, we have shown here that our \texttt{AugSVF} augmentation method dramatically reduces a model's susceptibility in these regimes. This gets at a broader question to be investigated: how reliable is the Fourier perspective on robustness? In particular, does improved robustness to plane wave perturbations in a particular subdomain of frequencies and orientations reliably indicate that a model will be more robust to corruptions that are prominently comprised of these plane wave components? Lastly, the simplicity and composability of Fourier bases may make them an especially effective augmentation strategy in applied domains, such as medical imaging, where other common photography-inspired augmentations are not appropriate; thus Fourier-based augmentations should be tested on a broad range of domain-specific datasets. 

\printbibliography

@inproceedings{cubuk2018autoaugment,
  title={Autoaugment Learning augmentation policies from data},
  author={Cubuk, Zoph, Mane and Vasudevan, Le},
  booktitle={Proceedings of the IEEE Conference on Computer Vision and Pattern Recognition},
  pages={113--123},
  year={2019}
}

@article{ford2019adversarial,
  title={Adversarial examples are a natural consequence of test error in noise},
  author={Ford, Nic and Gilmer, Justin and Carlini, Nicolas and Cubuk, Dogus},
  journal={arXiv preprint arXiv:1901.10513},
  year={2019}
}

@inproceedings{globerson2006nightmare,
  title={Nightmare at test time: robust learning by feature deletion},
  author={Globerson, Amir and Roweis, Sam},
  booktitle={Proceedings of the 23rd international conference on Machine learning},
  pages={353--360},
  year={2006}
}

@article{geirhos2018generalisation,
  title={Generalisation in humans and deep neural networks},
  author={Geirhos, Robert and Temme, Carlos RM and Rauber, Jonas and Sch{\"u}tt, Heiko H and Bethge, Matthias and Wichmann, Felix A},
  journal={Advances in Neural Information Processing Systems},
  volume={31},
  pages={7538--7550},
  year={2018}
}

@article{he2021towards,
  title={Towards non-iid image classification: A dataset and baselines},
  author={He, Yue and Shen, Zheyan and Cui, Peng},
  journal={Pattern Recognition},
  volume={110},
  pages={107383},
  year={2021},
  publisher={Elsevier}
}

@inproceedings{hendrycks2018benchmarking,
  title={Benchmarking Neural Network Robustness to Common Corruptions and Perturbations},
  author={Hendrycks, Dan and Dietterich, Thomas},
  booktitle={International Conference on Learning Representations},
  year={2018}
}

@inproceedings{hendrycks2019augmix,
  title={AugMix: A Simple Data Processing Method to Improve Robustness and Uncertainty},
  author={Hendrycks, Dan and Mu, Norman and Cubuk, Ekin Dogus and Zoph, Barret and Gilmer, Justin and Lakshminarayanan, Balaji},
  booktitle={International Conference on Learning Representations},
  year={2019}
}

@inproceedings{hendrycks2021natural,
  title={Natural adversarial examples},
  author={Hendrycks, Dan and Zhao, Kevin and Basart, Steven and Steinhardt, Jacob and Song, Dawn},
  booktitle={Proceedings of the IEEE/CVF Conference on Computer Vision and Pattern Recognition},
  pages={15262--15271},
  year={2021}
}

@inproceedings{jacobsen2018excessive,
  title={Excessive Invariance Causes Adversarial Vulnerability},
  author={Jacobsen, Joern-Henrik and Behrmann, Jens and Zemel, Richard and Bethge, Matthias},
  booktitle={International Conference on Learning Representations},
  year={2018}
}

@article{jo2017measuring,
  title={Measuring the tendency of cnns to learn surface statistical regularities},
  author={Jo, Jason and Bengio, Yoshua},
  journal={arXiv preprint arXiv:1711.11561},
  year={2017}
}

@inproceedings{kumar2019verified,
  title={Verified uncertainty calibration},
  author={Kumar, Ananya and Liang, Percy and Ma, Tengyu},
  booktitle={Proceedings of the 33rd International Conference on Neural Information Processing Systems},
  pages={3792--3803},
  year={2019}
}

@inproceedings{lu2020harder,
  title={Harder or different? a closer look at distribution shift in dataset reproduction},
  author={Lu, Shangyun and Nott, Bradley and Olson, Aaron and Todeschini, Alberto and Vahabi, Hossein and Carmon, Yair and Schmidt, Ludwig},
  booktitle={ICML Workshop on Uncertainty and Robustness in Deep Learning},
  year={2020}
}

@inproceedings{moosavi2017universal,
  title={Universal adversarial perturbations},
  author={Moosavi-Dezfooli, Seyed-Mohsen and Fawzi, Alhussein and Fawzi, Omar and Frossard, Pascal},
  booktitle={Proceedings of the IEEE conference on computer vision and pattern recognition},
  pages={1765--1773},
  year={2017}
}

@article{mu2019mnist,
  title={Mnist-c: A robustness benchmark for computer vision},
  author={Mu, Norman and Gilmer, Justin},
  journal={arXiv preprint arXiv:1906.02337},
  year={2019}
}

@inbook{newman_2013, place={Michigan}, 
   edition={Revised and Expanded}, 
   title={The Discrete Fourier Transform}, 
   booktitle={Computational Physics}, 
   publisher={Mark Newman}, 
   author={Newman, Mark}, 
   year={2013}, 
   pages={299–300}}

@article{recht2018cifar,
  title={Do cifar-10 classifiers generalize to cifar-10?},
  author={Recht, Benjamin and Roelofs, Rebecca and Schmidt, Ludwig and Shankar, Vaishaal},
  journal={arXiv preprint arXiv:1806.00451},
  year={2018}
}

@inproceedings{recht2019imagenet,
  title={Do imagenet classifiers generalize to imagenet?},
  author={Recht, Benjamin and Roelofs, Rebecca and Schmidt, Ludwig and Shankar, Vaishaal},
  booktitle={International Conference on Machine Learning},
  pages={5389--5400},
  year={2019},
  organization={PMLR}
}

@software{soklaski2021zen,
  author       = {Ryan Soklaski and
                  Justin Goodwin},
  title        = {{mit-ll-responsible-ai/hydra-zen: Release hydra-zen 
                   v0.3.0rc2}},
  month        = sep,
  year         = 2021,
  publisher    = {Zenodo},
  version      = {v0.3.0rc2},
  doi          = {10.5281/zenodo.5517572},
  url          = {https://doi.org/10.5281/zenodo.5517572}
}

@inproceedings{tsuzuku2019structural,
  title={On the Structural Sensitivity of Deep Convolutional Networks to the Directions of Fourier Basis Functions},
  author={Tsuzuku, Yusuke and Sato, Issei},
  booktitle={2019 IEEE/CVF Conference on Computer Vision and Pattern Recognition (CVPR)},
  pages={51--60},
  year={2019},
  organization={IEEE Computer Society}
}

@article{yin2019fourier,
  title={A fourier perspective on model robustness in computer vision},
  author={Yin, Dong and Lopes, Raphael Gontijo and Shlens, Jonathon and Cubuk, Ekin D and Gilmer, Justin},
  journal={arXiv preprint arXiv:1906.08988},
  year={2019}
}

@inproceedings{zagoruyko2016wide,
  title={Wide Residual Networks},
  author={Zagoruyko, Sergey and Komodakis, Nikos},
  booktitle={British Machine Vision Conference 2016},
  year={2016},
  organization={British Machine Vision Association}
}

@misc{wiki:jsdloss,
   author = "{Wikipedia contributors}",
   title = "Jensen–Shannon divergence --- {W}ikipedia{,} The Free Encyclopedia",
   year = "2004",
   url = "https://en.wikipedia.org/wiki/Jensen–Shannon_divergence",
   note = "[Online; accessed 10-November-2021]"
 }
\medskip



\appendix

\section{Appendix}

\subsection{Additional Implementation Details of Fourier-Based Perturbations}

Although our formulation of $U(\vec{k})$ presented in Section \ref{sec:fourier}, which explicitly defines a real-valued plane wave sinusoid, does not immediately resemble the Fourier-based definition introduced in \cite{yin2019fourier}, the two are exactly equivalent, as are their implementations.

A channel-aligned Fourier-basis perturbation is applied to an image by adding $U$, which is scaled to have a specified $\ell_2$-norm in order to control the strength of the perturbation, to each of the image's color channels. The image's pixel values are assumed to reside in $[0,1]$ and thus they are ``clipped'' to this domain following the channel-wise addition.

A random-flip perturbation follows this same process, but includes a randomly-drawn factor from $\{-1, 1\}$ that multiplies $U$ in association with each color channel. We include both channel-aligned and random-flip perturbations in our analyses, whereas the random-flip type is exclusively considered in some prior works \cite{hendrycks2019augmix,yin2019fourier}. 

\subsection{Additional Implementations Details of \texttt{AugMix}}

The \texttt{AugMix} data processing technique processes an image through parallel chains of randomly-selected and composed augmentation primitives; it randomly weights and sums these parallel-processed images, and then randomly mixes the result with the original image \cite{hendrycks2019augmix}.

An example involving a particular \texttt{AugMix} configuration, for \texttt{AugSVF}, for processing a single image is detailed in equation 
\ref{eq:augmix}.


\begin{equation}
\arraycolsep=1.8pt
\def\arraystretch{1.2}
\begin{array}{rcll}
(w_1, w_2, w_3) & \sim & \mathrm{Dirichlet}(1,1,1) & \\
m & \sim & \mathrm{Beta}(1,1) & \\
\mathrm{AugSVF}(x_\mathrm{img}) & = & (1-m) \; x_\mathrm{img} \; + \; m \; & \big[ w_1 \; p_{\mathrm{rotate}} \circ p_{\mathrm{shear_{x}}}(x_\mathrm{img}) \\
& & & + \; w_2 \; p_{\mathrm{EQ}}(x_\mathrm{img}) \\
& & & + \; w_3 \; p_{\mathrm{fourier}} \circ p_{\mathrm{rotate}}(x_\mathrm{img}) \big]
\end{array}
\label{eq:augmix}
\end{equation}

Here the ``depth''---the number of augmentation primitives composed for each chain---of each of the three chains is drawn uniformly from $\{1, 2, 3\}$ with replacement. Once a chain's depth has been determined, that number of augmentations is drawn uniformly from $S \cup V \cup F$. The example in \ref{eq:augmix} thus shows depths of 2, 1, and 2, respectively.

Prior to this \texttt{AugMix} processing, a random flip-and-crop is applied to the image---as in the ``baseline'' augmentation strategy. Following the \texttt{AugMix} process, the per-channel data normalization---enforcing a per-channel mean and standard deviation of 0.5---is applied to the ``mixed'' image.

\subsection{Methods}

Our Wide ResNet architecture configuration, hyperparameters, and training methods match exactly those used in \cite{hendrycks2019augmix} except for one detail: for each seed, we select the best model checkpoint based on performance on a holdout validation set, whereas the cited work evaluates directly on the test set. We leveraged hydra-zen \cite{soklaski2021zen} to produce all of our results in a self-documenting and reproducible manner.

Each Fourier-basis perturbation used for the purpose of train-time augmentation has an $\ell_2$-norm sampled uniformly from $[1, 2]$. The frequency and orientation are sampled in a weighted fashion such that the distinct frequencies are uniformly represented. Lastly, the phase shift is sampled uniformly from $\{\frac{0}{3}2\pi, \frac{1}{3}2\pi, \frac{2}{3}2\pi\}$. All test-time evaluations involving Fourier-basis perturbations (e.g., generating a heatmap), utilize a distinct phase shift of $\frac{3}{4}2\pi$.

\subsection{Additional Results}

\begin{figure}[!htb]
  \centering
  \includegraphics[width=5.5in]{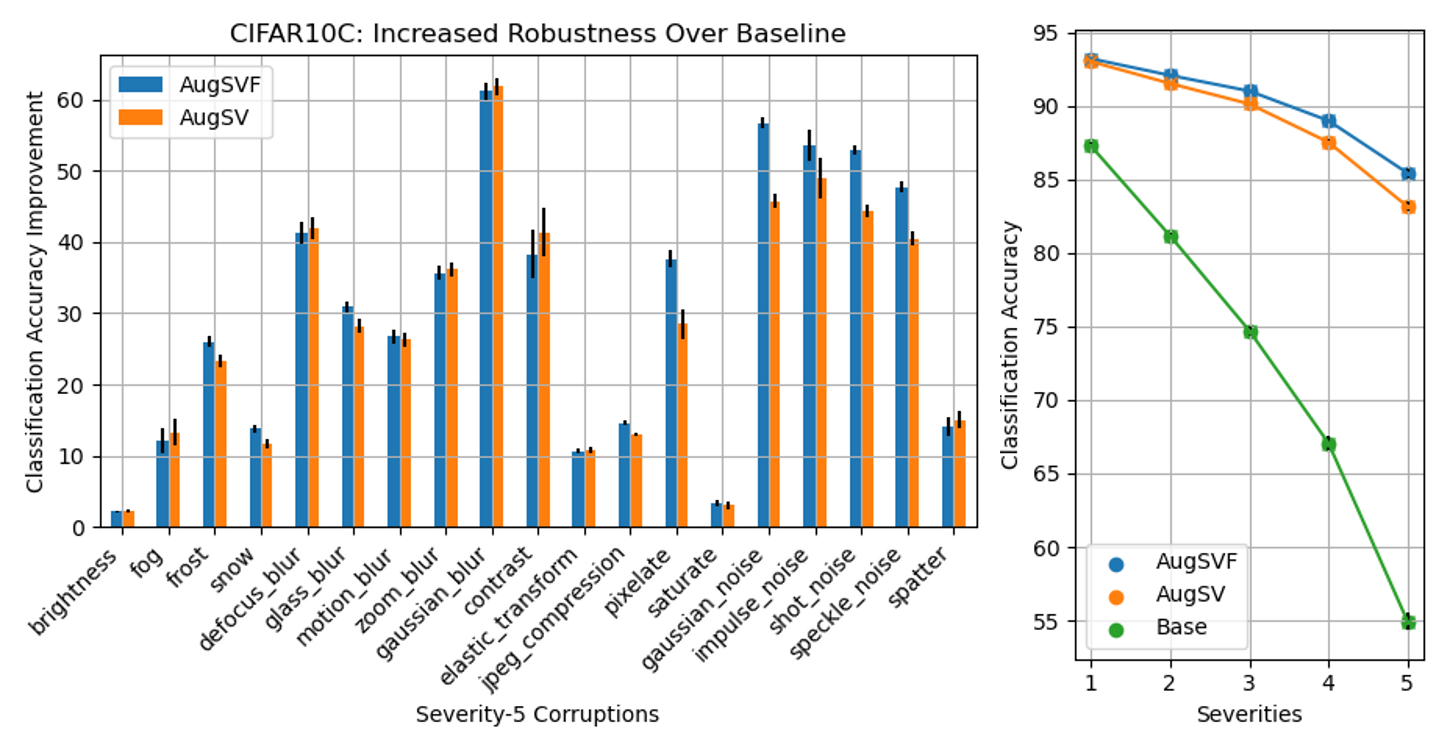}
  \caption{(Left) Enhanced robustness in classification accuracy over the baseline, across severity-5 corruption categories for CIFAR-10-C. \texttt{AugSVF} yields a significant boost in robustness to ``noise'' and ``digital'' corruptions, without any significant trade-offs elsewhere. (Right) CIFAR-10-C classification accuracies averaged over corruptions, for each corruption severity.}
\label{fig:c10_bars}
\end{figure}

Figure \ref{fig:c10_bars} is the complement to Figure \ref{fig:c100_bars}: it reports the gains in robustness---due to \texttt{AugSV} and \texttt{AugSVF}, respectively---for the various corruption categories in CIFAR-10-C.

\subsubsection{Further Presentation and Analysis of Susceptibility Heatmaps}

Here we include heatmaps corresponding to Fourier attacks of various channel alignments and strengths (i.e. $\norm{U}$) against the \texttt{Base}, \texttt{AugSV}, and \texttt{AugSVF} models, which were each trained separately on CIFAR-10 and CIFAR-100. Note the dramatic reduction in susceptibility that is achieved by \texttt{AugSVF} for all channel alignments and strengths. It should be noted that perturbations with $\norm{U} = 4$ are roughly comparable to severity-5 corruptions from the CIFAR-C datasets, in terms of how visually disruptive they are.

It is interesting to see the different susceptibility patterns that manifest, between the channel-aligned and random-flip attacks; the former tend to have stronger effects in the high-frequency regimes, while the latter impact mid-to-low frequency regimes more severely. Prior papers \cite{hendrycks2019augmix,yin2019fourier} only considered random-flip channels, which overlooked an important class of Fourier perturbations and their impact on model performance. For instance \texttt{AugSV} appears to be somewhat robust to Fourier perturbations, in comparison to \texttt{Base}, when only considering random-flip attacks. However, e.g, for CIFAR-10, we see that high error-rates (80\%) manifest when \texttt{AugSV} encounters high-frequency, channel-aligned perturbations that are only of strength $\norm{U} = 2$.

\begin{figure}[!htb]
  \centering
  \includegraphics[width=5.5in]{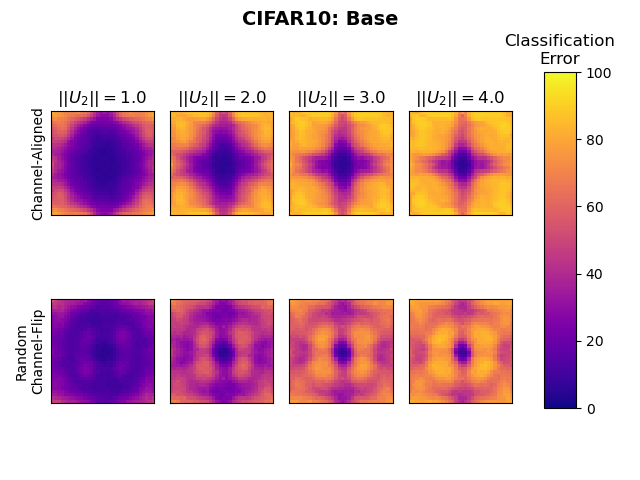}
\end{figure}

\begin{figure}[!htb]
  \centering
  \includegraphics[width=5.5in]{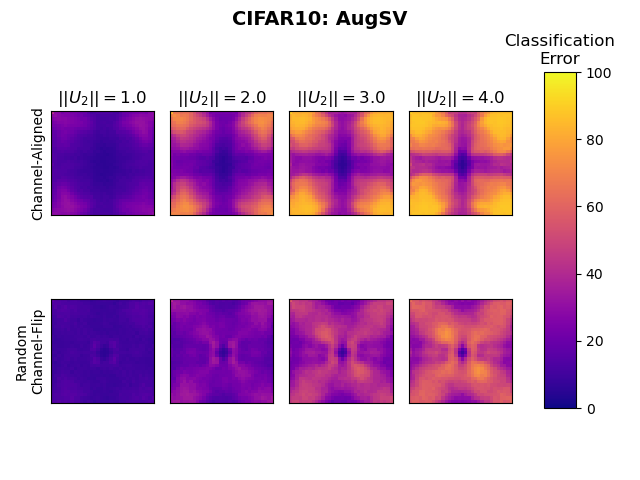}
\end{figure}

\begin{figure}[!htb]
  \centering
  \includegraphics[width=5.5in]{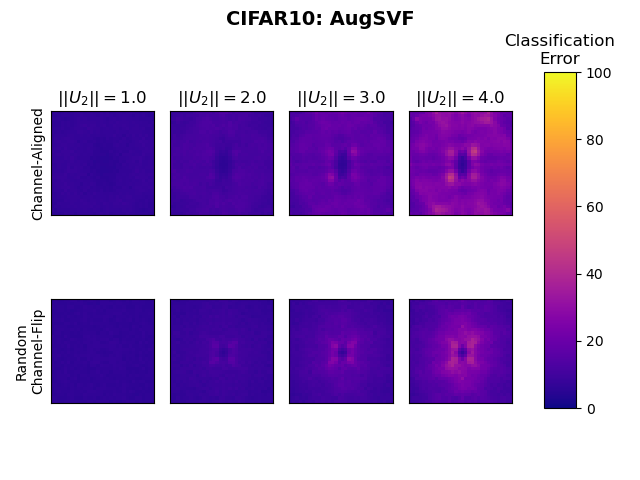}
\end{figure}

\begin{figure}[!htb]
  \centering
  \includegraphics[width=5.5in]{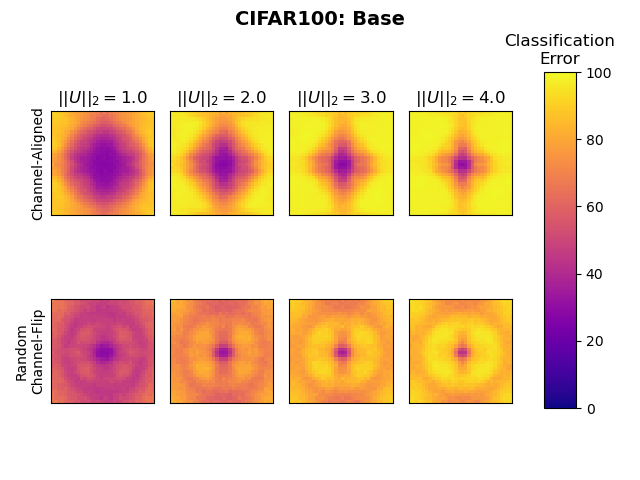}
\end{figure}

\begin{figure}[!htb]
  \centering
  \includegraphics[width=5.5in]{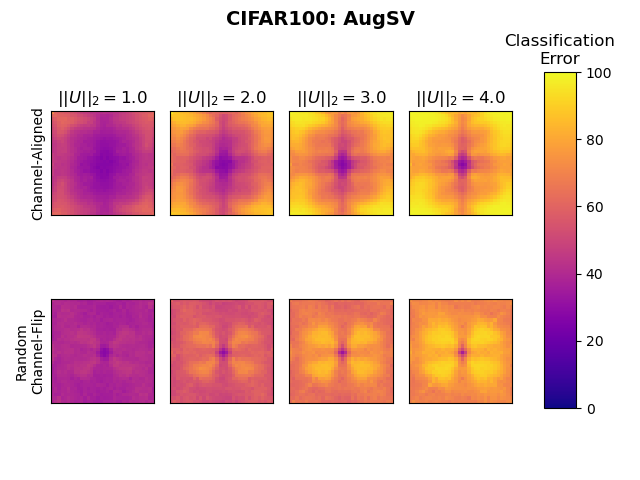}
\end{figure}

\begin{figure}[!htb]
  \centering
  \includegraphics[width=5.5in]{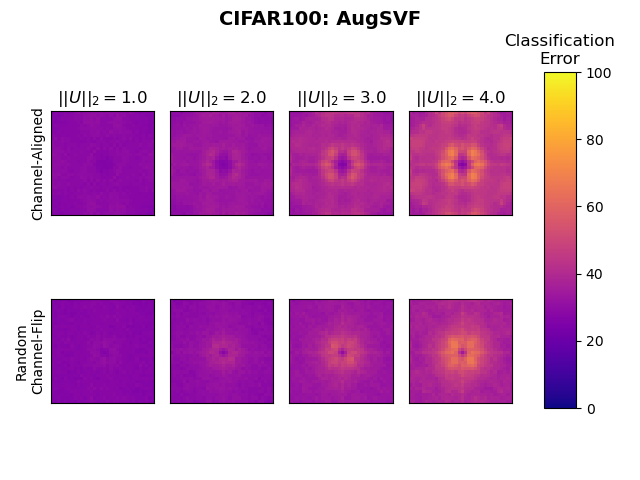}
\end{figure}

\end{document}